\documentclass[10pt,twocolumn,letterpaper]{article}

\usepackage{colortbl}
\usepackage{cvpr}
\usepackage{times}
\usepackage{epsfig}
\usepackage{graphicx}
\usepackage{amsmath}
\usepackage{amssymb}

\usepackage{microtype}

\usepackage{booktabs}
\usepackage[table]{xcolor}

\usepackage[breaklinks=true,letterpaper=true,colorlinks,bookmarks=false]{hyperref}

\cvprfinalcopy 

\def\cvprPaperID{2315} 

\ifcvprfinal\fi

\definecolor{yelloworange}{RGB}{255, 153, 0}
\definecolor{ultramarineblue}{RGB}{65, 102, 245}

\newcommand{\bm}[1]{\boldsymbol{#1}}

\newtheorem{lemma-ap}{Lemma}

\newtheorem{claim-ap}{Claim}

\makeatletter
\usepackage{xspace}
\def\@onedot{\ifx\@let@token.\else.\null\fi\xspace}
\DeclareRobustCommand\onedot{\futurelet\@let@token\@onedot}

\newcommand{\figref}[1]{Fig\onedot~\ref{#1}}

\newcommand{\secref}[1]{Sec\onedot~\ref{#1}}
\newcommand{\tabref}[1]{Tab\onedot~\ref{#1}}

\def\eg{\emph{e.g}\onedot} 
\def\ie{\emph{i.e}\onedot} 
\def\cf{\emph{cf}\onedot} 
 \def\vs{\emph{vs}\onedot}
 
\def\etal{\emph{et al}\onedot}

\newcommand{\eat}[1]{}

\makeatother

\begin{document}

\title{Rethinking Atrous Convolution for Semantic Image Segmentation}

\author{
  \begin{tabular}[t]{c c c c}
    Liang-Chieh~Chen & George~Papandreou & Florian~Schroff & Hartwig~Adam \\
    \multicolumn{4}{c}{Google Inc.} \\
    \multicolumn{4}{c}{\{lcchen, gpapan, fschroff, hadam\}@google.com} \\
\end{tabular}
}

\maketitle

\begin{abstract}
  In this work, we revisit atrous convolution, a powerful tool to explicitly adjust filter's field-of-view
  as well as control the resolution of feature responses computed by Deep Convolutional Neural Networks,
  in the application of semantic image segmentation. To handle the problem of segmenting
  objects at multiple scales, we design modules which employ atrous convolution in
  cascade or in parallel to capture multi-scale context by adopting multiple atrous rates.
  Furthermore, we propose to augment our previously proposed Atrous Spatial Pyramid Pooling module, which probes
  convolutional features at multiple scales, with image-level features
  encoding global context and further boost performance. We also elaborate on implementation details and
  share our experience on training our system. The proposed `DeepLabv3' system significantly
  improves over our previous DeepLab versions without DenseCRF post-processing and attains comparable performance with
  other state-of-art models on the PASCAL VOC 2012 semantic image segmentation benchmark.
\end{abstract}

\section{Introduction}
For the task of semantic segmentation \cite{everingham2014pascal,mottaghi2014role,Cordts2016Cityscapes,zhou2017scene, caesar2016cocostuff}, we consider two challenges in applying Deep Convolutional Neural Networks (DCNNs) \cite{lecun1989backpropagation}. The first one is the reduced feature resolution caused by consecutive pooling operations or convolution striding, which allows DCNNs to learn increasingly abstract feature representations. However, this invariance to local image transformation may impede dense prediction tasks, where detailed spatial information is desired. To overcome this problem, we advocate the use of atrous convolution \cite{holschneider1989real, giusti2013fast, sermanet2013overfeat, papandreou2014untangling}, which has been shown to be effective for semantic image segmentation \cite{chen2014semantic, yu2015multi, chen2016deeplab}. Atrous convolution, also known as dilated convolution, allows us to repurpose ImageNet \cite{ILSVRC15} pretrained networks to extract denser feature maps by removing the downsampling operations from the last few layers and upsampling the corresponding filter kernels, equivalent to inserting holes (`trous' in French) between filter weights. With atrous convolution, one is able to control the resolution at which feature responses are computed within DCNNs without requiring learning extra parameters. 

\begin{figure}[!t]
  \centering
  \begin{tabular}{c}
    \includegraphics[width=0.95\linewidth]{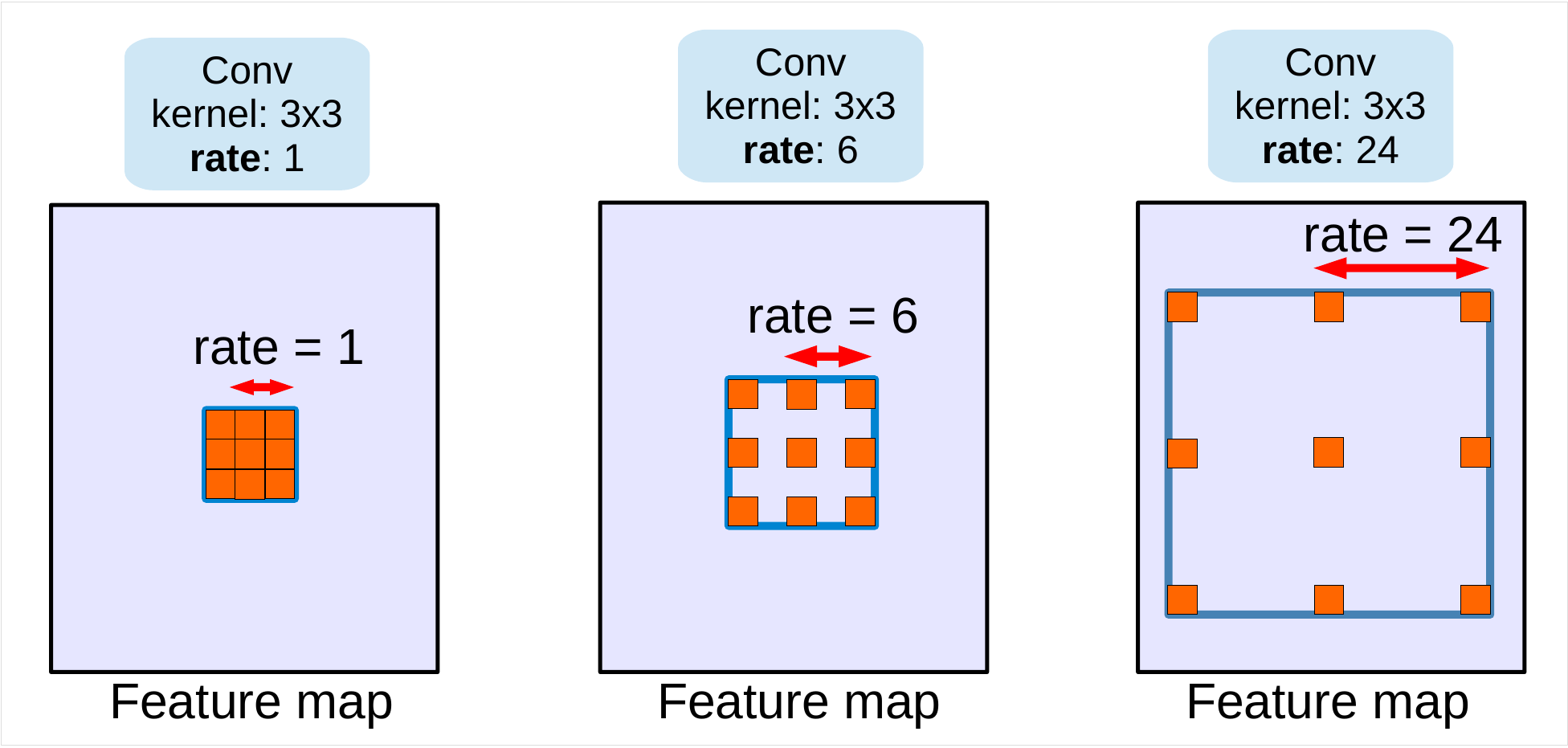} \\
  \end{tabular}
  \caption{Atrous convolution with kernel size $3 \times 3$ and different rates. Standard convolution corresponds to atrous convolution with $rate=1$. Employing large value of atrous rate enlarges the model's field-of-view, enabling object encoding at multiple scales.}
  \label{fig:atrous}
\end{figure}

\begin{figure*}[!t]
  \centering
  \begin{tabular}{c c c c}
    \includegraphics[height=0.22\linewidth]{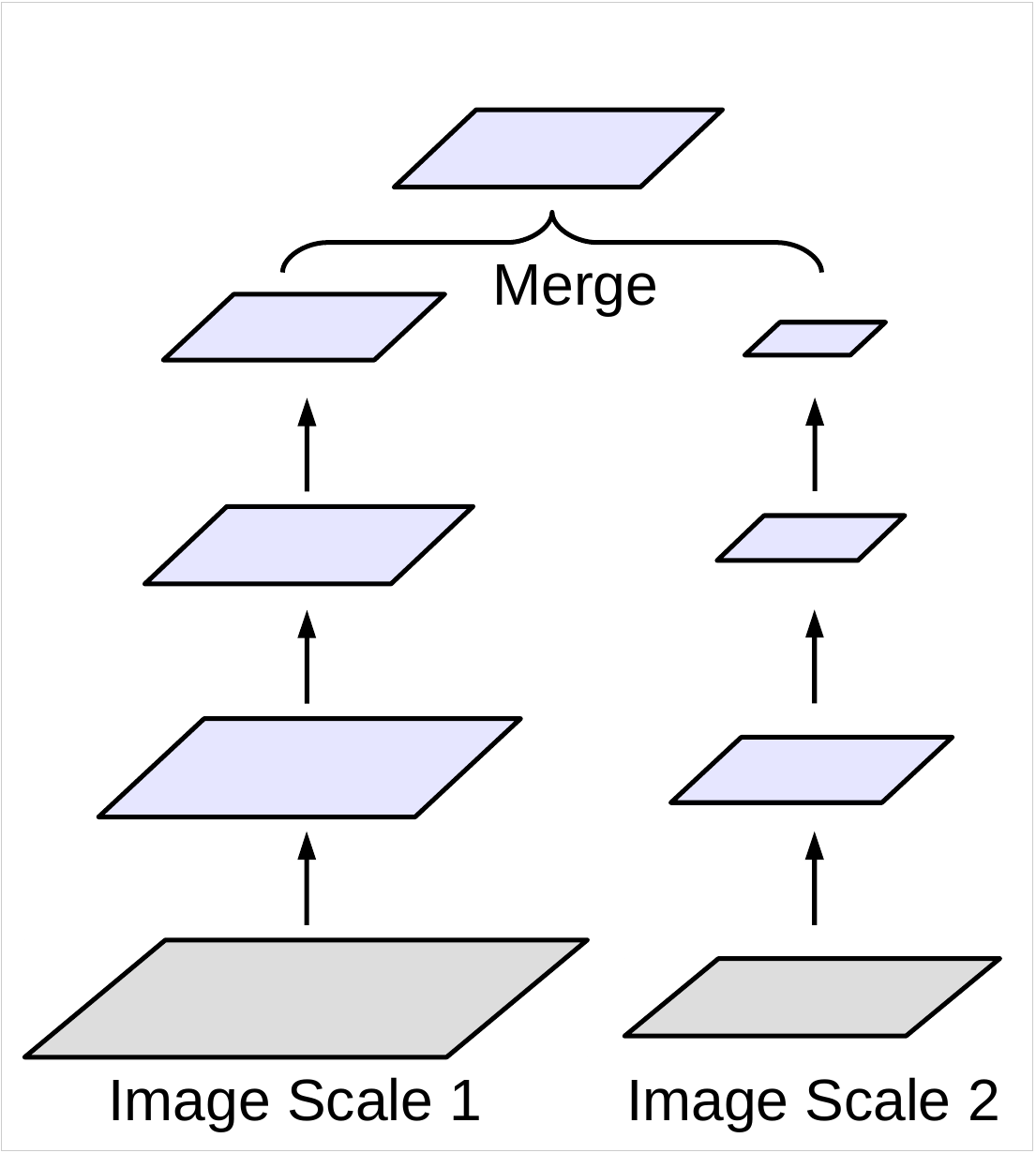} &
    \includegraphics[height=0.22\linewidth]{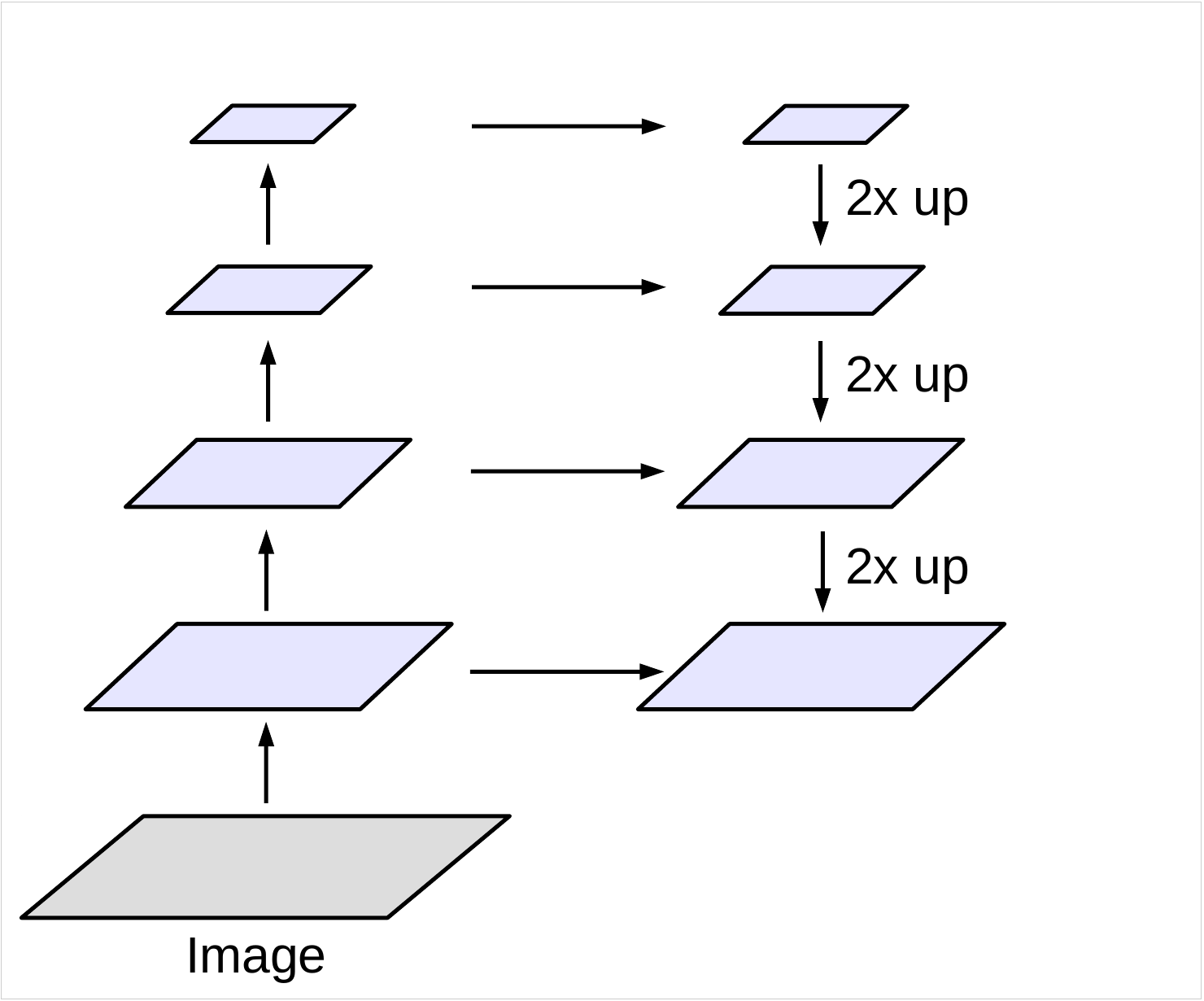} &
    \includegraphics[height=0.22\linewidth]{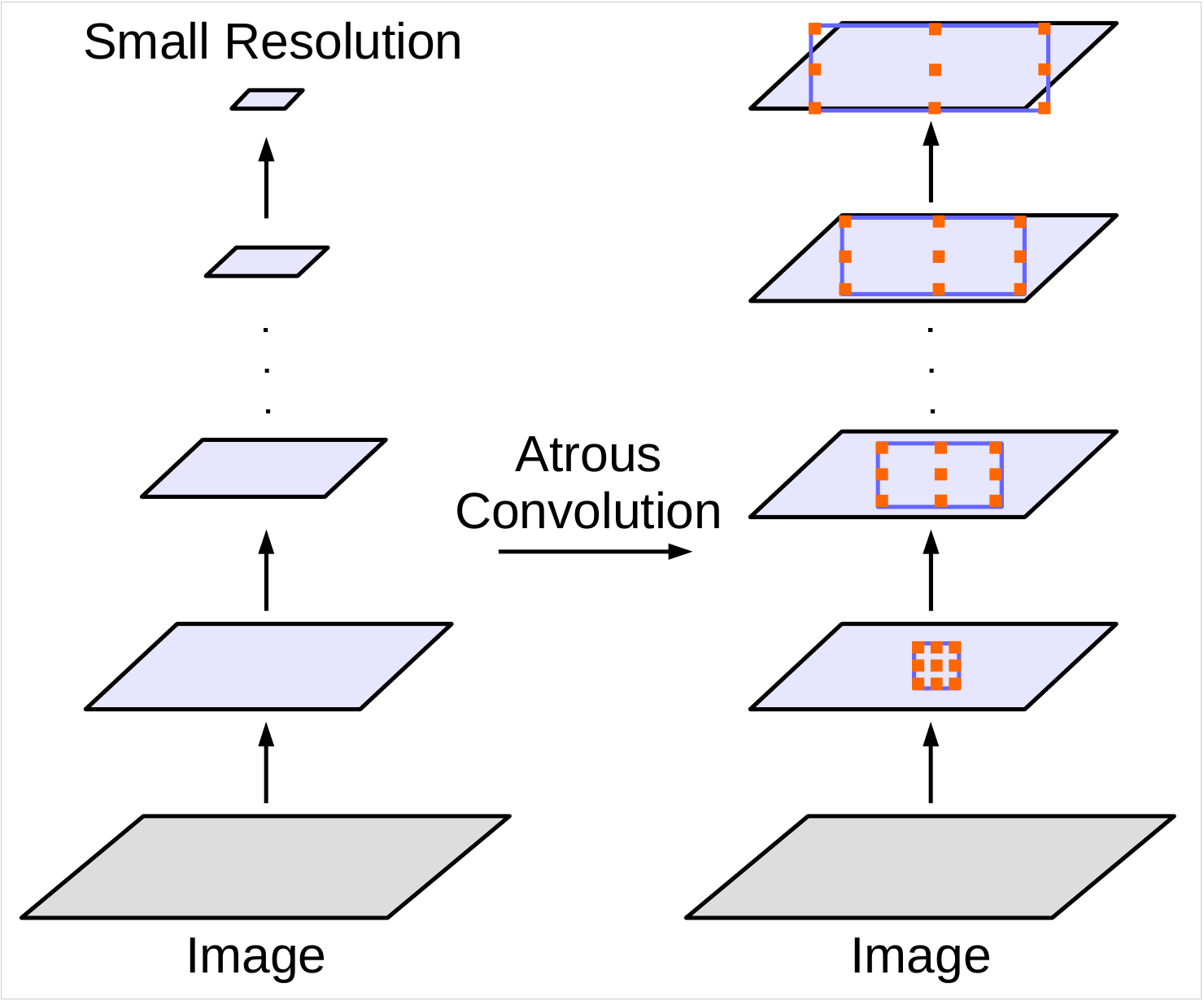} &
    \includegraphics[height=0.22\linewidth]{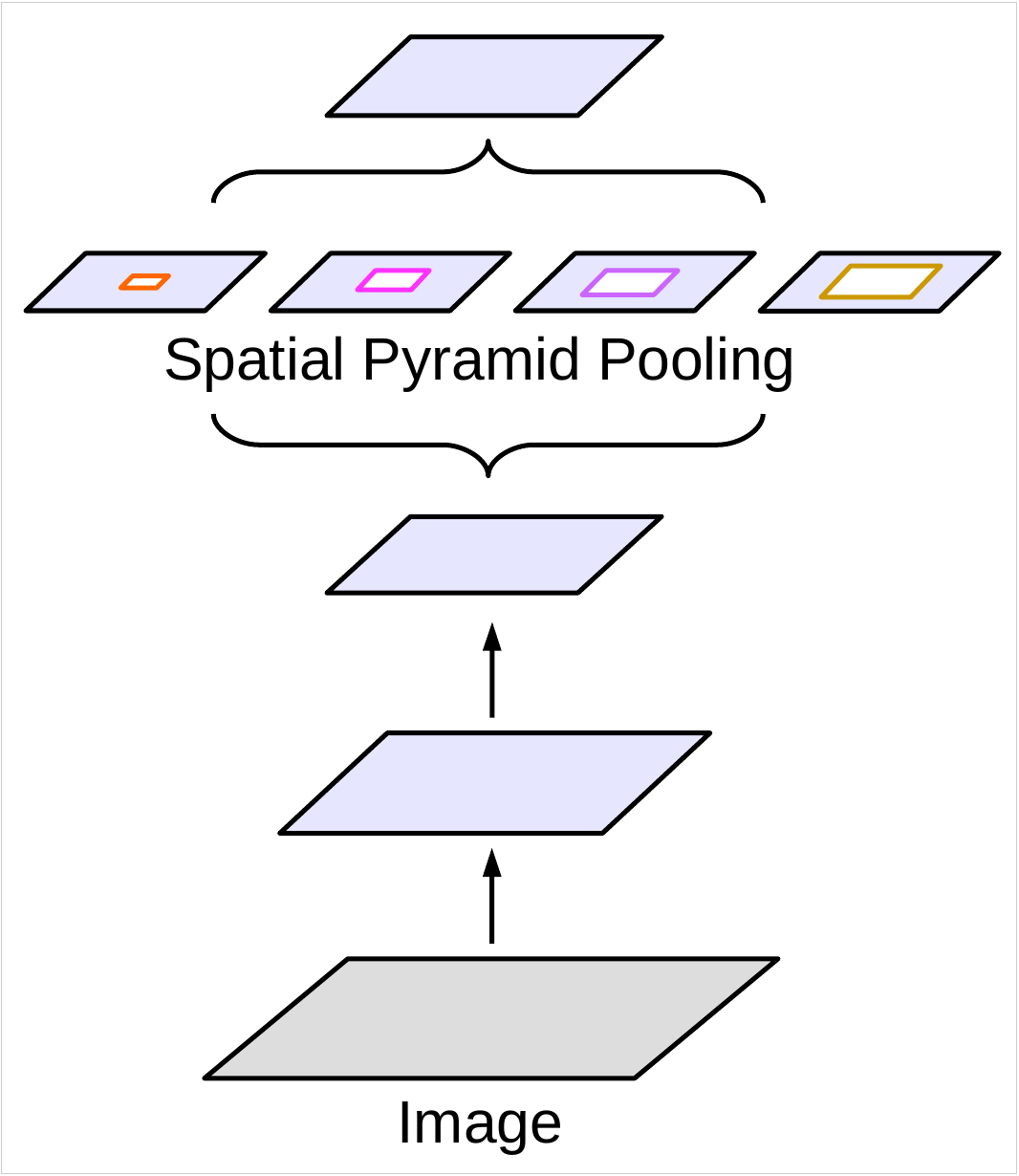} \\
    \small{(a) Image Pyramid} &
    \small{(b) Encoder-Decoder} &
    \small{(c) Deeper w. Atrous Convolution} &
    \small{(d) Spatial Pyramid Pooling} \\
  \end{tabular}
  \caption{Alternative architectures to capture multi-scale context.}
  \label{fig:arch}
\end{figure*}

Another difficulty comes from the existence of objects at multiple scales. Several methods have been proposed to handle the problem and we mainly consider four categories in this work, as illustrated in \figref{fig:arch}. First, the DCNN is applied to an image pyramid to extract features for each scale input \cite{farabet2013learning, eigen2014predicting, pinheiro2014recurrent, lin2015efficient,chen2015attention,chen2016deeplab} where objects at different scales become prominent at different feature maps. Second, the encoder-decoder structure \cite{badrinarayanan2015segnet,ronneberger2015u,ghiasi2016laplacian,lin2016refinenet,pohlen2016full, peng2017large, islamgated} exploits multi-scale features from the encoder part and recovers the spatial resolution from the decoder part. Third, extra modules are cascaded on top of the original network for capturing long range information. In particular, DenseCRF \cite{krahenbuhl2011efficient} is employed to encode pixel-level pairwise similarities \cite{chen2014semantic, zheng2015conditional, lin2015efficient, schwing2015fully}, while \cite{liu2015semantic, yu2015multi} develop several extra convolutional layers in cascade to gradually capture long range context. Fourth, spatial pyramid pooling \cite{chen2016deeplab, zhao2016pyramid} probes an incoming feature map with filters or pooling operations at multiple rates and multiple effective field-of-views, thus capturing objects at multiple scales.

In this work, we revisit applying atrous convolution, which allows us to effectively enlarge the field of view of filters to incorporate multi-scale context, in the framework of both cascaded modules and spatial pyramid pooling. In particular, our proposed module consists of atrous convolution with various rates and batch normalization layers which we found important to be trained as well. We experiment with laying out the modules in cascade or in parallel (specifically, Atrous Spatial Pyramid Pooling (ASPP) method \cite{chen2016deeplab}). We discuss an important practical issue when applying a $3\times3$ atrous convolution with an extremely large rate, which fails to capture long range information due to image boundary effects, effectively simply degenerating to $1\times1$ convolution, and propose to incorporate image-level features into the ASPP module. Furthermore, we elaborate on implementation details and share experience on training the proposed models, including a simple yet effective bootstrapping method for handling rare and finely annotated objects. In the end, our proposed model, `DeepLabv3' improves over our previous works \cite{chen2014semantic, chen2016deeplab} and attains performance of 85.7\% on the PASCAL VOC 2012 \textit{test} set without DenseCRF post-processing. 

\section{Related Work}
It has been shown that global features or contextual interactions \cite{he2004multiscale,shotton2009textonboost,kohli2009robust,ladicky2009associative,gould2009decomposing,yao2012describing} are beneficial in correctly classifying pixels for semantic segmentation. In this work, we discuss four types of Fully Convolutional Networks (FCNs) \cite{sermanet2013overfeat, long2014fully} (see \figref{fig:arch} for illustration) that exploit context information for semantic segmentation \cite{hariharan2014hypercolumns, dai2014convolutional, mostajabi2014feedforward, chen2015semantic, zheng2015conditional, lin2015efficient, papandreou2015weakly, schwing2015fully, xia2015zoom}.

\textbf{Image pyramid:} The same model, typically with shared weights, is applied to multi-scale inputs. Feature responses from the small scale inputs encode the long-range context, while the large scale inputs preserve the small object details. Typical examples include Farabet \etal \cite{farabet2013learning} who transform the input image through a Laplacian pyramid, feed each scale input to a DCNN and merge the feature maps from all the scales. \cite{eigen2014predicting, pinheiro2014recurrent} apply multi-scale inputs sequentially from coarse-to-fine, while \cite{lin2015efficient,chen2015attention,chen2016deeplab} directly resize the input for several scales and fuse the features from all the scales. The main drawback of this type of models is that it does not scale well for larger/deeper DCNNs (\eg, networks like \cite{he2015deep,zagoruyko2016wide,wu2016wider}) due to limited GPU memory and thus it is usually applied during the inference stage \cite{dai2015boxsup}.

\textbf{Encoder-decoder:} This model consists of two parts: (a) the encoder where the spatial dimension of feature maps is gradually reduced and thus longer range information is more easily captured in the deeper encoder output, and (b) the decoder where object details and spatial dimension are gradually recovered. For example, \cite{long2014fully,noh2015learning} employ deconvolution \cite{zeiler2011adaptive} to learn the upsampling of low resolution feature responses. SegNet \cite{badrinarayanan2015segnet} reuses the pooling indices from the encoder and learn extra convolutional layers to densify the feature responses, while U-Net \cite{ronneberger2015u} adds skip connections from the encoder features to the corresponding decoder activations, and \cite{ghiasi2016laplacian} employs a Laplacian pyramid reconstruction network. More recently, RefineNet \cite{lin2016refinenet} and \cite{pohlen2016full, peng2017large, islamgated} have demonstrated the effectiveness of models based on encoder-decoder structure on several semantic segmentation benchmarks. This type of model is also explored in the context of object detection \cite{lin2016feature,shrivastava2016beyond}.

\textbf{Context module:} This model contains extra modules laid out in cascade to encode long-range context. One effective method is to incorporate DenseCRF \cite{krahenbuhl2011efficient} (with efficient high-dimensional filtering algorithms \cite{adams2010fast}) to DCNNs \cite{chen2014semantic, chen2016deeplab}. Furthermore, \cite{zheng2015conditional, lin2015efficient, schwing2015fully} propose to jointly train both the CRF and DCNN components, while \cite{liu2015semantic, yu2015multi} employ several extra convolutional layers on top of the belief maps of DCNNs (belief maps are the final DCNN feature maps that contain output channels equal to the number of predicted classes) to capture context information. Recently, \cite{jampani2016learning} proposes to learn a general and sparse high-dimensional convolution (bilateral convolution), and \cite{Vemulapalli2016Gaussian, chandra2016fast} combine Gaussian Conditional Random Fields and DCNNs for semantic segmentation.

\textbf{Spatial pyramid pooling:} This model employs spatial pyramid pooling \cite{grauman2005pyramid, lazebnik2006beyond} to capture context at several ranges. The image-level features are exploited in ParseNet \cite{liu2015parsenet} for global context information. DeepLabv2 \cite{chen2016deeplab} proposes atrous spatial pyramid pooling (ASPP), where parallel atrous convolution layers with different rates capture multi-scale information. Recently, Pyramid Scene Parsing Net (PSP) \cite{zhao2016pyramid} performs spatial pooling at several grid scales and demonstrates outstanding performance on several semantic segmentation benchmarks. There are other methods based on LSTM \cite{hochreiter1997long} to aggregate global context \cite{liang2015semantic, byeon2015scene, yan2016combining}. Spatial pyramid pooling has also been applied in object detection \cite{he2014spatial}.

In this work, we mainly explore atrous convolution \cite{holschneider1989real, giusti2013fast, sermanet2013overfeat, papandreou2014untangling, chen2014semantic, yu2015multi, chen2016deeplab} as a {\bf{context module}} and tool for {\bf{spatial pyramid pooling}}. Our proposed framework is general in the sense that it could be applied to any network. To be concrete, we duplicate several copies of the original last block in ResNet \cite{he2015deep} and arrange them in cascade, and also revisit the ASPP module \cite{chen2016deeplab} which contains several atrous convolutions in parallel. Note that our cascaded modules are applied directly on the feature maps instead of belief maps. For the proposed modules, we experimentally find it important to train with batch normalization \cite{ioffe2015batch}. To further capture global context, we propose to augment ASPP with image-level features, similar to \cite{liu2015parsenet, zhao2016pyramid}.

\textbf{Atrous convolution:} Models based on atrous convolution have been actively explored for semantic segmentation. For example, \cite{wu2016bridging} experiments with the effect of modifying atrous rates for capturing long-range information, \cite{wang2017understanding} adopts hybrid atrous rates within the last two blocks of ResNet, while \cite{dai2017deformable} further proposes to learn the deformable convolution which samples the input features with learned offset, generalizing atrous convolution. To further improve the segmentation model accuracy, \cite{wanglearning} exploits image captions, \cite{jain2017fusionseg} utilizes video motion, and \cite{kong2017recurrent} incorporates depth information. Besides, atrous convolution has been applied to object detection by \cite{papandreou2014untangling, dai2016rfcn, huang2016speed}.

\section{Methods}
\label{sec:methods}
In this section, we review how atrous convolution is applied to extract dense features for semantic segmentation. We then discuss the proposed modules with atrous convolution modules employed in cascade or in parallel.

\begin{figure*}[!t]
  \centering
  \begin{tabular}{c}
    \includegraphics[width=0.95\linewidth]{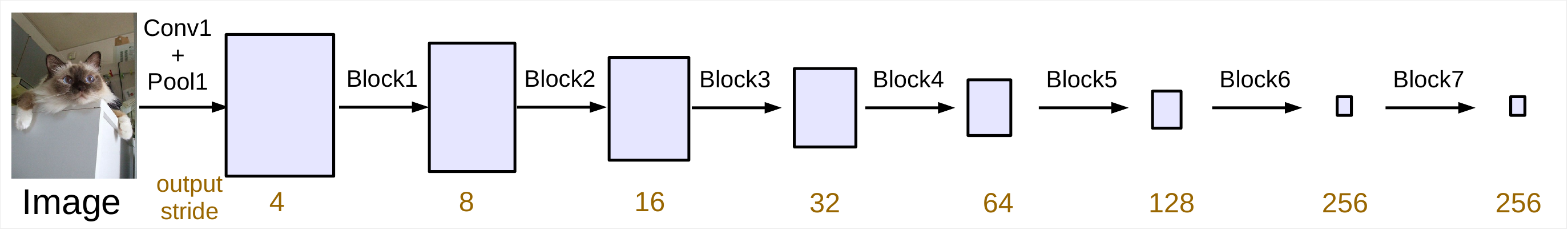} \\
    \small{(a) Going deeper without atrous convolution.}\\
    \includegraphics[width=0.95\linewidth]{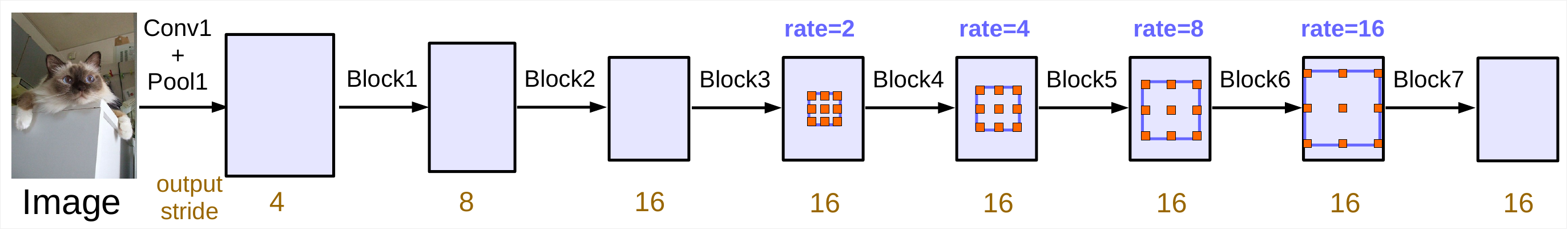} \\
    \small{(b) Going deeper with atrous convolution. Atrous convolution with $rate > 1$ is applied after block3 when $\emph{output\_stride}=16$.}\\
  \end{tabular}
  \caption{Cascaded modules without and with atrous convolution.}
  \label{fig:model_deeper}
\end{figure*}

\subsection{Atrous Convolution for Dense Feature Extraction}
Deep Convolutional Neural Networks (DCNNs) \cite{lecun1989backpropagation} deployed in fully convolutional fashion \cite{sermanet2013overfeat, long2014fully} have shown to be effective for the task of semantic segmentation. However, the repeated combination of max-pooling and striding at consecutive layers of these networks significantly reduces the spatial resolution of the resulting feature maps, typically by a factor of 32 across each direction in recent DCNNs \cite{krizhevsky2012imagenet, simonyan2014very, he2015deep}. Deconvolutional layers (or transposed convolution) \cite{zeiler2011adaptive, long2014fully, noh2015learning, badrinarayanan2015segnet, ronneberger2015u, peng2017large} have been employed to recover the spatial resolution. Instead, we advocate the use of `atrous convolution', originally developed for the efficient computation of the undecimated wavelet transform in the ``algorithme \`a trous'' scheme of \cite{holschneider1989real} and used before in the DCNN context by \cite{giusti2013fast, sermanet2013overfeat, papandreou2014untangling}. 

Consider two-dimensional signals, for each location $\bm{i}$ on the output $\bm{y}$ and a filter $\bm{w}$, atrous convolution is applied over the input feature map $\bm{x}$:

\begin{equation}
  \bm{y}[\bm{i}] = \sum_{\bm{k}} \bm{x}[\bm{i} + r \cdot \bm{k}] \bm{w}[\bm{k}]
\end{equation}
where
the atrous rate \emph{r} corresponds to the stride with which we sample the input signal, which is equivalent to convolving the input $\bm{x}$ with upsampled filters produced by inserting $r - 1$ zeros between two consecutive filter values along each spatial dimension (hence the name atrous convolution where the French word trous means holes in English). Standard convolution is a special case for rate $r = 1$, and atrous convolution allows us to adaptively modify filter's field-of-view by changing the rate value. See \figref{fig:atrous} for illustration.

Atrous convolution also allows us to explicitly control how densely to compute feature responses in fully convolutional networks. Here, we denote by \emph{output\_stride} the ratio of input image spatial resolution to final output resolution. For the DCNNs \cite{krizhevsky2012imagenet, simonyan2014very, he2015deep} deployed for the task of image classification, the final feature responses (before fully connected layers or global pooling) is 32 times smaller than the input image dimension, and thus $\emph{output\_stride}=32$. If one would like to double the spatial density of computed feature responses in the DCNNs (\ie, $\emph{output\_stride} = 16$), the stride of last pooling or convolutional layer that decreases resolution is set to 1 to avoid signal decimation. Then, all subsequent convolutional layers are replaced with atrous convolutional layers having rate $r = 2$. This allows us to extract denser feature responses without requiring learning any extra parameters. Please refer to \cite{chen2016deeplab} for more details.

\subsection{Going Deeper with Atrous Convolution}
We first explore designing modules with atrous convolution laid out in cascade. To be concrete, we duplicate several copies of the last ResNet block, denoted as block4 in \figref{fig:model_deeper}, and arrange them in cascade. There are three $3\times3$ convolutions in those blocks, and the last convolution contains stride 2 except the one in last block, similar to original ResNet. The motivation behind this model is that the introduced striding makes it easy to capture long range information in the deeper blocks. For example, the whole image feature could be summarized in the last small resolution feature map, as illustrated in \figref{fig:model_deeper}~(a). However, we discover that the consecutive striding is harmful for semantic segmentation (see \tabref{tab:deeper_res50} in \secref{sec:experiments}) since detail information is decimated, and thus we apply atrous convolution with rates determined by the desired \emph{output\_stride} value, as shown in \figref{fig:model_deeper}~(b) where $\emph{output\_stride}=16$.

In this proposed model, we experiment with cascaded ResNet blocks up to block7 (\ie, extra block5, block6, block7 as replicas of block4), which has $\emph{output\_stride}=256$ if no atrous convolution is applied.

\subsubsection{Multi-grid Method}
Motivated by multi-grid methods which employ a hierarchy of grids of different sizes \cite{brandt1977multi, terzopoulos1986image, briggs2000multigrid, papandreou2007multigrid} and following \cite{wang2017understanding, dai2017deformable}, we adopt different atrous rates within block4 to block7 in the proposed model. In particular, we define as $\emph{Multi\_Grid}=(r_1,r_2,r_3)$ the unit rates for the three convolutional layers within block4 to block7. The final atrous rate for the convolutional layer is equal to the multiplication of the unit rate and the corresponding rate. For example, when $\emph{output\_stride}=16$ and $\emph{Multi\_Grid}=(1, 2, 4)$, the three convolutions will have $rates=2 \cdot (1, 2, 4) = (2, 4, 8)$ in the block4, respectively.

\subsection{Atrous Spatial Pyramid Pooling}
We revisit the Atrous Spatial Pyramid Pooling proposed in \cite{chen2016deeplab}, where four parallel atrous convolutions with different atrous rates are applied on top of the feature map. ASPP is inspired by the success of spatial pyramid pooling \cite{grauman2005pyramid, lazebnik2006beyond, he2014spatial} which showed that it is effective to resample features at different scales for accurately and efficiently classifying regions of an arbitrary scale. Different from \cite{chen2016deeplab}, we include batch normalization within ASPP. 

\begin{figure}[!t]
  \centering
  \begin{tabular}{c}
    \includegraphics[width=0.7\linewidth]{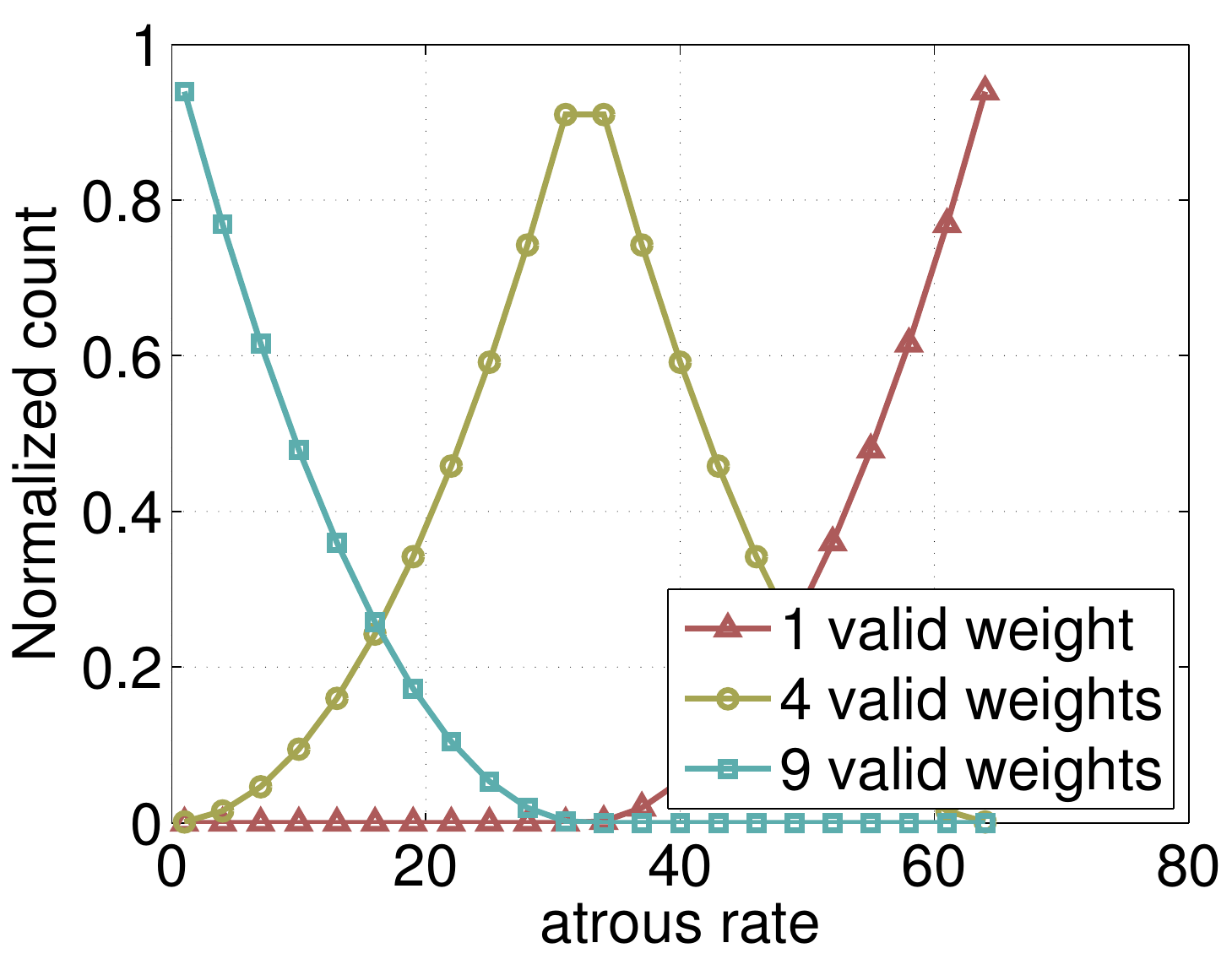}\\
  \end{tabular}
  \caption{Normalized counts of valid weights with a $3\times3$ filter on a $65\times65$ feature map as atrous rate varies. When atrous rate is small, all the 9 filter weights are applied to most of the valid region on feature map, while atrous rate gets larger, the $3\times3$ filter degenerates to a $1\times1$ filter since only the center weight is effective.}
  \label{fig:effective_region}
\end{figure}

ASPP with different atrous rates effectively captures multi-scale information. However, we discover that as the sampling rate becomes larger, the number of valid filter weights (\ie, the weights that are applied to the valid feature region, instead of padded zeros) becomes smaller. This effect is illustrated in \figref{fig:effective_region} when applying a $3\times3$ filter to a $65\times65$ feature map with different atrous rates. In the extreme case where the rate value is close to the feature map size, the $3\times3$ filter, instead of capturing the whole image context, degenerates to a simple $1\times1$ filter since only the center filter weight is effective.

\begin{figure*}[!t]
  \centering
  \begin{tabular}{c}
    \includegraphics[width=0.95\linewidth]{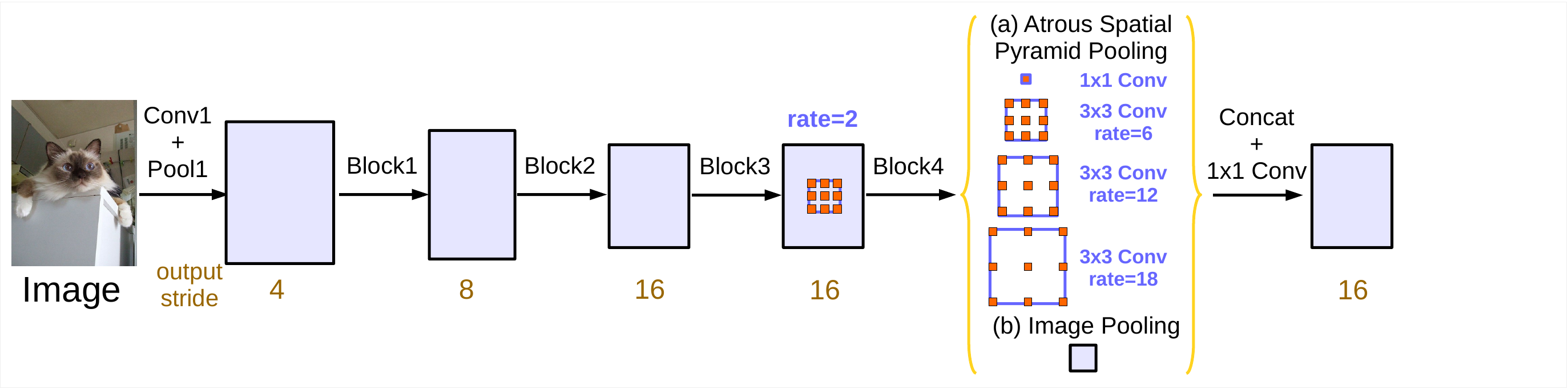} \\
  \end{tabular}
  \caption{Parallel modules with atrous convolution (ASPP), augmented with image-level features.}
  \label{fig:model_aspp}
\end{figure*}

To overcome this problem and incorporate global context information to the model, we adopt image-level features, similar to \cite{liu2015parsenet, zhao2016pyramid}. Specifically, we apply global average pooling on the last feature map of the model, feed the resulting image-level features to a $1\times1$ convolution with 256 filters (and batch normalization \cite{ioffe2015batch}), and then bilinearly upsample the feature to the desired spatial dimension. In the end, our improved ASPP consists of (a) one $1\times1$ convolution and three $3\times3$ convolutions with $rates=(6, 12, 18)$ when $\emph{output\_stride}=16$ (all with 256 filters and batch normalization), and (b) the image-level features, as shown in \figref{fig:model_aspp}. Note that the rates are doubled when $\emph{output\_stride}=8$. The resulting features from all the branches are then concatenated and pass through another $1\times1$ convolution (also with 256 filters and batch normalization) before the final $1\times1$ convolution which generates the final logits.

\section{Experimental Evaluation}
\label{sec:experiments}
We adapt the ImageNet-pretrained \cite{ILSVRC15} ResNet \cite{he2015deep} to the semantic segmentation by applying atrous convolution to extract dense features. Recall that \emph{output\_stride} is defined as the ratio of input image spatial resolution to final output resolution. For example, when $\emph{output\_stride}=8$, the last two blocks (block3 and block4 in our notation) in the original ResNet contains atrous convolution with $rate=2$ and $rate=4$ respectively. Our implementation is built on TensorFlow \cite{abadi2016tensorflow}.

We evaluate the proposed models on the PASCAL VOC 2012 semantic segmentation benchmark \cite{everingham2014pascal} which contains 20 foreground object classes and one background class. The original dataset contains $1,464$ (\textit{train}), $1,449$ (\textit{val}), and $1,456$ (\textit{test}) pixel-level labeled
images for training, validation, and testing, respectively. The dataset is
augmented by the extra annotations provided by \cite{hariharan2011semantic},
resulting in $10,582$ (\textit{trainaug}) training images. The performance
is measured in terms of pixel intersection-over-union (IOU) averaged across
the 21 classes.

\subsection{Training Protocol}
\label{subsec:train_protocol}
In this subsection, we discuss details of our training protocol.

\textbf{Learning rate policy:} Similar to \cite{liu2015parsenet, chen2016deeplab}, we employ a ``poly'' learning rate policy where the initial learning rate is multiplied by $(1-\frac{iter}{max\_iter})^{power}$ with $power = 0.9$.

\textbf{Crop size:} Following the original training protocol \cite{chen2014semantic, chen2016deeplab}, patches are cropped from the image during training. For atrous convolution with large rates to be effective, large crop size is required; otherwise, the filter weights with large atrous rate are mostly applied to the padded zero region. We thus employ crop size to be 513 during both training and test on PASCAL VOC 2012 dataset.

\textbf{Batch normalization:} Our added modules on top of ResNet all include batch normalization parameters \cite{ioffe2015batch}, which we found important to be trained as well. Since large batch size is required to train batch normalization parameters, we employ $\emph{output\_stride}=16$ and compute the batch normalization statistics with a batch size of 16.
The batch normalization parameters are trained with decay = 0.9997. After training on the \textit{trainaug} set with 30K iterations and initial learning rate = 0.007, we then freeze batch normalization parameters, employ $\emph{output\_stride}=8$, and train on the official PASCAL VOC 2012 \textit{trainval} set for another 30K iterations and smaller base learning rate = 0.001. Note that atrous convolution allows us to control \emph{output\_stride} value at different training stages without requiring learning extra model parameters. Also note that training with $\emph{output\_stride}=16$ is several times faster than $\emph{output\_stride}=8$ since the intermediate feature maps are spatially four times smaller, but at a sacrifice of accuracy since $\emph{output\_stride}=16$ provides coarser feature maps.

\textbf{Upsampling logits:} In our previous works \cite{chen2014semantic, chen2016deeplab}, the target groundtruths are downsampled by 8 during training when $\emph{output\_stride}=8$. We find it important to keep the groundtruths intact and instead upsample the final logits, since downsampling the groundtruths removes the fine annotations resulting in no back-propagation of details.

\textbf{Data augmentation:} We apply data augmentation by randomly scaling the input images (from 0.5 to 2.0) and randomly left-right flipping during training.

\subsection{Going Deeper with Atrous Convolution}
We first experiment with building more blocks with atrous convolution in cascade.

\textbf{ResNet-50:} In \tabref{tab:deeper_res50}, we experiment with the effect of \emph{output\_stride} when employing ResNet-50 with block7 (\ie, extra block5, block6, and block7). As shown in the table, in the case of $\emph{output\_stride}=256$ (\ie, no atrous convolution at all), the performance is much worse than the others due to the severe signal decimation. When \emph{output\_stride} gets larger and apply atrous convolution correspondingly, the performance improves from 20.29\% to 75.18\%, showing that atrous convolution is essential when building more blocks cascadedly for semantic segmentation.

\begin{table}[!t]
  \centering
  \scalebox{0.85}{
  \begin{tabular}{c | c c c c c c}
    \toprule[0.2em]
    \emph{output\_stride} & 8 & 16 & 32 & 64 & 128 & 256 \\
    \toprule[0.2em]
    mIOU & 75.18 & 73.88 & 70.06 & 59.99 & 42.34 & 20.29 \\
    \bottomrule[0.1em]
  \end{tabular}
  }
  \caption{Going deeper with atrous convolution when employing ResNet-50 with block7 and different \emph{output\_stride}. Adopting $\emph{output\_stride}=8$ leads to better performance at the cost of more memory usage.}
  \label{tab:deeper_res50}
\end{table}

\begin{table}[!t]
  \centering
  \scalebox{1.}{
  \begin{tabular}{c | c c c c}
    \toprule[0.2em]
    Network & block4 & block5 & block6 & block7 \\
    \toprule[0.2em]
    ResNet-50 & 64.81 & 72.14 & 74.29 & 73.88 \\
    ResNet-101 & 68.39 & 73.21 & 75.34 & 75.76 \\
    \bottomrule[0.1em]
  \end{tabular}
  }
  \caption{Going deeper with atrous convolution when employing ResNet-50 and ResNet-101 with different number of cascaded blocks at $\emph{output\_stride}=16$. Network structures `block4', `block5', `block6', and `block7' add extra 0, 1, 2, 3 cascaded modules respectively. The performance is generally improved by adopting more cascaded blocks.}
  \label{tab:deeper_res101}
\end{table}

\begin{table}[!t]
  \centering
  \scalebox{1.}{
  \begin{tabular}{c | c c c c}
    \toprule[0.2em]
    Multi-Grid & block4 & block5 & block6 & block7 \\
    \toprule[0.2em]
    (1, 1, 1) & 68.39 & 73.21 & 75.34 & 75.76 \\
    (1, 2, 1) & 70.23 & 75.67 & 76.09 & {\bf 76.66} \\
    (1, 2, 3) & 73.14 & 75.78 & 75.96 & 76.11 \\
    (1, 2, 4) & 73.45 & 75.74 & 75.85 & 76.02 \\
    (2, 2, 2) & 71.45 & 74.30 & 74.70 & 74.62 \\
    \bottomrule[0.1em]
  \end{tabular}
  }
  \caption{Employing multi-grid method for ResNet-101 with different number of cascaded blocks at $\emph{output\_stride}=16$. The best model performance is shown in bold.}
  \label{tab:multigrid}
\end{table}

\textbf{ResNet-50 vs. ResNet-101:} We replace ResNet-50 with deeper network ResNet-101 and change the number of cascaded blocks. As shown in \tabref{tab:deeper_res101}, the performance improves as more blocks are added, but the margin of improvement becomes smaller. Noticeably, employing block7 to ResNet-50 decreases slightly the performance while it still improves the performance for ResNet-101.

\textbf{Multi-grid:} We apply the multi-grid method to ResNet-101 with several cascadedly added blocks in \tabref{tab:multigrid}. The unit rates, $\emph{Multi\_Grid}=(r_1,r_2,r_3)$, are applied to block4 and all the other added blocks. As shown in the table, we observe that (a) applying multi-grid method is generally better than the vanilla version where $(r_1,r_2,r_3)=(1, 1, 1)$, (b) simply doubling the unit rates (\ie, $(r_1,r_2,r_3)=(2, 2, 2)$) is not effective, and (c) going deeper with multi-grid improves the performance. Our best model is the case where block7 and $(r_1,r_2,r_3)=(1,2,1)$ are employed. 

\textbf{Inference strategy on val set:} The proposed model is trained with $\emph{output\_stride}=16$, and then during inference we apply $\emph{output\_stride}=8$ to get more detailed feature map. As shown in \tabref{tab:deeper_val}, interestingly, when evaluating our best cascaded model with $\emph{output\_stride}=8$, the performance improves over evaluating with $\emph{output\_stride}=16$ by $1.39\%$. The performance is further improved by performing inference on multi-scale inputs (with $scales=\{0.5, 0.75, 1.0, 1.25, 1.5, 1.75\}$) and also left-right flipped images. In particular, we compute as the final result the average probabilities from each scale and flipped images.

\begin{table}[!t]
  \centering
  \scalebox{0.95}{
  \begin{tabular}{c|c c c c | c}
    \toprule[0.2em]
    Method & OS=16 & OS=8 & MS & Flip & mIOU \\
    \toprule[0.2em]
    block7 + & \checkmark &                 &    &      & 76.66 \\
    MG(1, 2, 1) &            & \checkmark      &    &      & 78.05 \\
    &            & \checkmark      & \checkmark & & 78.93 \\
    &            & \checkmark      & \checkmark & \checkmark & 79.35 \\
    \bottomrule[0.1em]
  \end{tabular}
  }
  \caption{Inference strategy on the \textit{val} set. {\bf MG}: Multi-grid. {\bf OS}: \emph{output\_stride}. {\bf MS}: Multi-scale inputs during test. {\bf Flip}: Adding left-right flipped inputs.}
  \label{tab:deeper_val}
\end{table}

\subsection{Atrous Spatial Pyramid Pooling}
We then experiment with the Atrous Spatial Pyramid Pooling (ASPP) module with the main differences from \cite{chen2016deeplab} being that batch normalization parameters \cite{ioffe2015batch} are fine-tuned and image-level features are included.

\begin{table}[!t]
  \centering
  \scalebox{0.7}{
  \begin{tabular}{c c c|c c | c| c }
    \toprule[0.2em]
    \multicolumn{3}{c|}{Multi-Grid} & \multicolumn{2}{c|}{ASPP} & Image & \\
    (1, 1, 1) & (1, 2, 1) & (1, 2, 4) & (6, 12, 18) & (6, 12, 18, 24) & Pooling & mIOU\\
    \toprule[0.2em]
    \checkmark &         &           & \checkmark &          &    & 75.36 \\
    & \checkmark &        & \checkmark &          &    & 75.93 \\
    &           & \checkmark & \checkmark &       &     & 76.58 \\
    &           & \checkmark &            & \checkmark  &   & 76.46 \\
    &           & \checkmark & \checkmark &  &  \checkmark & 77.21 \\
    \bottomrule[0.1em]
  \end{tabular}
  }
  \caption{Atrous Spatial Pyramid Pooling with multi-grid method and image-level features at $\emph{output\_stride}=16$.}
  \label{tab:aspp}
\end{table}

\begin{table}[!t]
  \centering
  \scalebox{0.8}{
  \begin{tabular}{c|c c c c c | c}
    \toprule[0.2em]
    Method & OS=16 & OS=8 & MS & Flip & COCO & mIOU \\
    \toprule[0.2em]
    MG(1, 2, 4) + & \checkmark &               &   &    &      & 77.21 \\
    ASPP(6, 12, 18) + & & \checkmark      &    &  &     & 78.51 \\
    Image Pooling & & \checkmark      & \checkmark & & & 79.45 \\
    & & \checkmark      & \checkmark & \checkmark & & 79.77 \\
    & & \checkmark      & \checkmark & \checkmark & \checkmark & 82.70 \\
    \bottomrule[0.1em]
  \end{tabular}
  }
  \caption{Inference strategy on the \textit{val} set: {\bf MG}: Multi-grid. {\bf ASPP}: Atrous spatial pyramid pooling. {\bf OS}: \emph{output\_stride}. {\bf MS}: Multi-scale inputs during test. {\bf Flip}: Adding left-right flipped inputs. {\bf COCO}: Model pretrained on MS-COCO.}
  \label{tab:aspp_val}
\end{table}

\textbf{ASPP:} In \tabref{tab:aspp}, we experiment with the effect of incorporating multi-grid in block4 and image-level features to the improved ASPP module. We first fix $ASPP=(6,12,18)$ (\ie, employ $rates=(6,12,18)$ for the three parallel $3\times3$ convolution branches), and vary the multi-grid value. Employing $\emph{Multi\_Grid}=(1,2,1)$ is better than $\emph{Multi\_Grid}=(1,1,1)$, while further improvement is attained by adopting $\emph{Multi\_Grid}=(1,2,4)$ in the context of $ASPP=(6,12,18)$ (\cf, the `block4' column in \tabref{tab:multigrid}). If we additionally employ another parallel branch with $rate = 24$ for longer range context, the performance drops slightly by 0.12\%. On the other hand, augmenting the ASPP module with image-level feature is effective, reaching the final performance of 77.21\%.

\textbf{Inference strategy on val set:} Similarly, we apply $\emph{output\_stride}=8$ during inference once the model is trained. As shown in \tabref{tab:aspp_val}, employing $\emph{output\_stride}=8$ brings 1.3\% improvement over using $\emph{output\_stride}=16$, adopting multi-scale inputs and adding left-right flipped images further improve the performance by 0.94\% and 0.32\%, respectively. The best model with ASPP attains the performance of 79.77\%, better than the best model with cascaded atrous convolution modules (79.35\%), and thus is selected as our final model for test set evaluation.

\textbf{Comparison with DeepLabv2:} Both our best cascaded model (in \tabref{tab:deeper_val}) and ASPP model (in \tabref{tab:aspp_val}) (in both cases without DenseCRF post-processing or MS-COCO pre-training) already outperform DeepLabv2 (77.69\% with DenseCRF and pretrained on MS-COCO in Tab. 4 of \cite{chen2016deeplab}) on the PASCAL VOC 2012 \textit{val} set. The improvement mainly comes from including and fine-tuning batch normalization parameters \cite{ioffe2015batch} in the proposed models and having a better way to encode multi-scale context.


\textbf{Appendix:} We show more experimental results, such as the effect of hyper parameters and Cityscapes \cite{Cordts2016Cityscapes} results, in the appendix.

\begin{figure*}[!th]
  \centering
  \begin{tabular}{c}
    \includegraphics[width=0.95\linewidth]{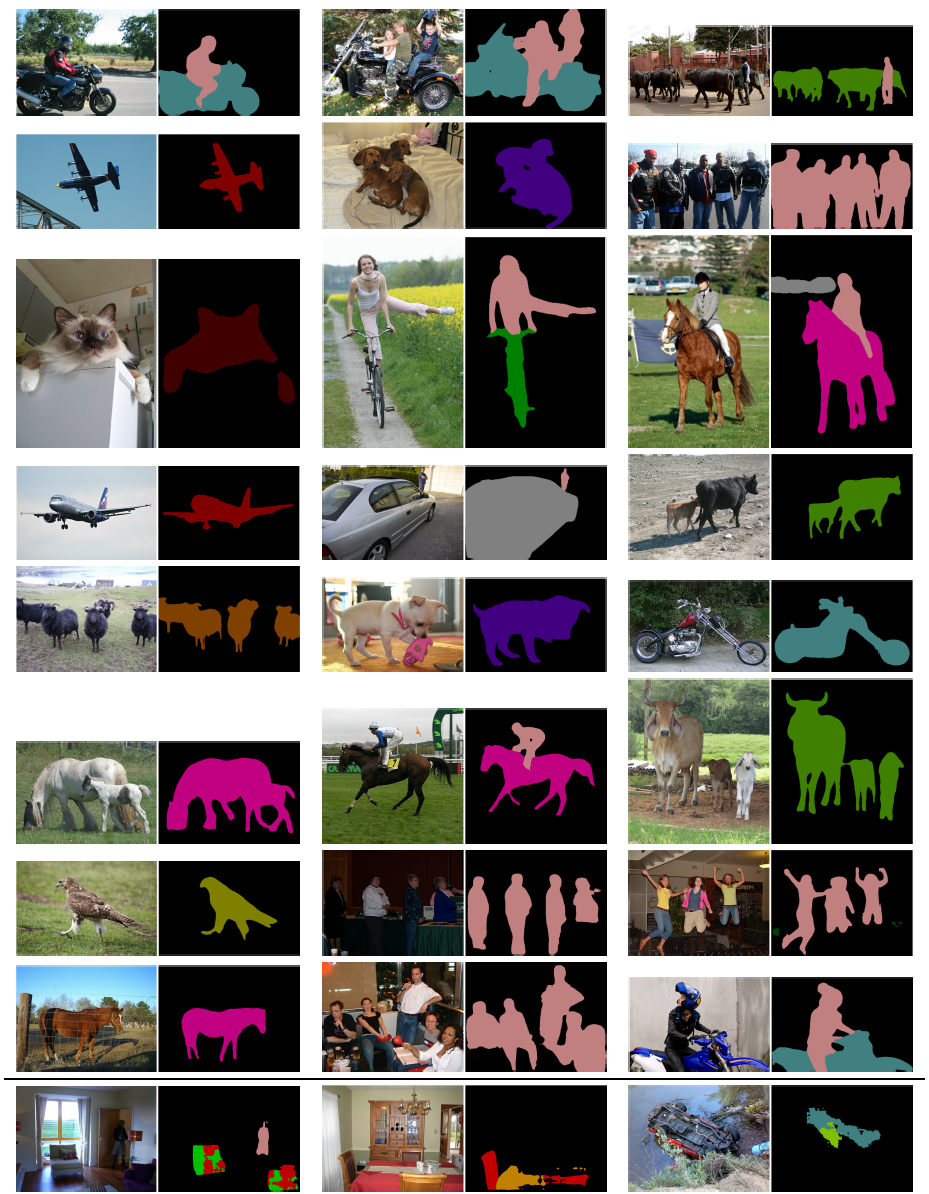}\\
  \end{tabular}
  \caption{Visualization results on the \textit{val} set when employing our best ASPP model. The last row shows a failure mode.}
  \label{fig:vis_results}
\end{figure*}

\textbf{Qualitative results:} We provide qualitative visual results of our best ASPP model in \figref{fig:vis_results}. As shown in the figure, our model is able to segment objects very well without any DenseCRF post-processing.

\textbf{Failure mode:} As shown in the bottom row of \figref{fig:vis_results}, our model has difficulty in segmenting (a) sofa \vs chair, (b) dining table and chair, and (c) rare view of objects.

\textbf{Pretrained on COCO:} For comparison with other state-of-art models, we further pretrain our best ASPP model on MS-COCO dataset \cite{lin2014microsoft}. From the MS-COCO \textit{trainval\_minus\_minival} set, we only select the images that have annotation regions larger than 1000 pixels and contain the classes defined in PASCAL VOC 2012, resulting in about 60K images for training. Besides, the MS-COCO classes not defined in PASCAL VOC 2012 are all treated as background class. After pretraining on MS-COCO dataset, our proposed model attains performance of 82.7\% on \textit{val} set when using $\emph{output\_stride}=8$, multi-scale inputs and adding left-right flipped images during inference. We adopt smaller initial learning rate = 0.0001 and same training protocol as in \secref{subsec:train_protocol} when fine-tuning on PASCAL VOC 2012 dataset.

\begin{figure}[!t]
  \centering
  \begin{tabular}{c c c c}
    \includegraphics[width=0.22\linewidth]{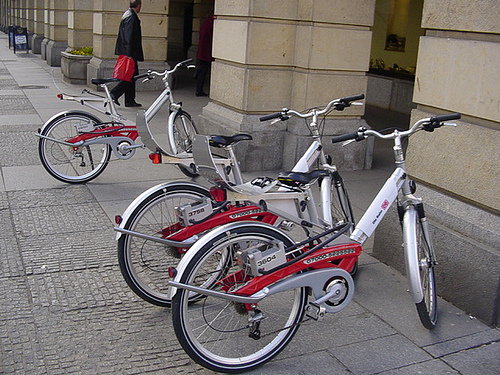} &
    \includegraphics[width=0.22\linewidth]{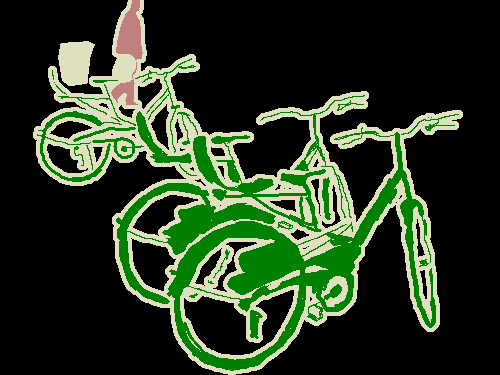} &
    \includegraphics[width=0.22\linewidth]{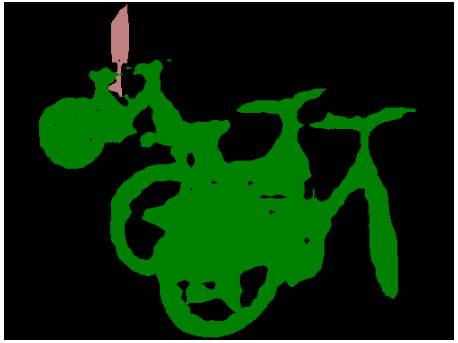} &
    \includegraphics[width=0.22\linewidth]{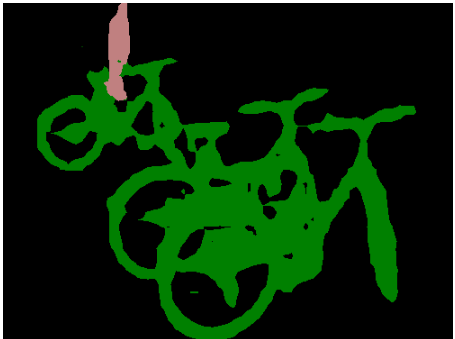} \\
    \includegraphics[width=0.22\linewidth]{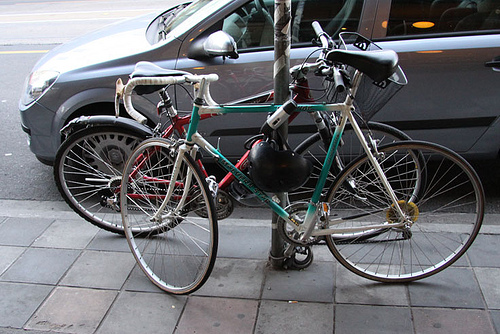} &
    \includegraphics[width=0.22\linewidth]{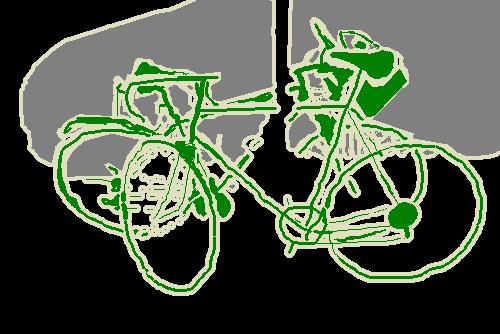} &
    \includegraphics[width=0.22\linewidth]{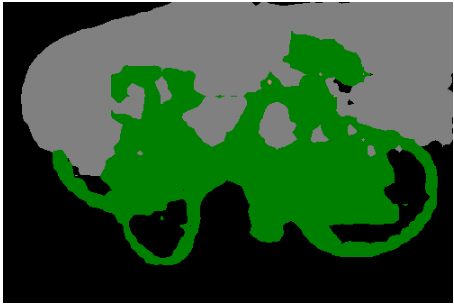} &
    \includegraphics[width=0.22\linewidth]{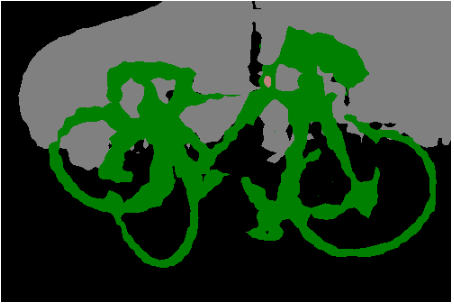} \\
    \tiny{(a) Image} &
    \tiny{(b) G.T.} &
    \tiny{(c) w/o bootstrapping} &
    \tiny{(d) w/ bootstrapping} \\
  \end{tabular}
  \caption{Bootstrapping on hard images improves segmentation accuracy for rare and finely annotated classes such as bicycle.}
  \label{fig:bicycle}
\end{figure}

\textbf{Test set result and an effective bootstrapping method:} We notice that PASCAL VOC 2012 dataset provides higher quality of annotations than the augmented dataset \cite{hariharan2011semantic}, especially for the bicycle class. We thus further fine-tune our model on the official PASCAL VOC 2012 \textit{trainval} set before evaluating on the test set. Specifically, our model is trained with $\emph{output\_stride}=8$ (so that annotation details are kept) and the batch normalization parameters are frozen (see \secref{subsec:train_protocol} for details).  Besides, instead of performing pixel hard example mining as \cite{wu2016bridging, pohlen2016full}, we resort to bootstrapping on hard images. In particular, we duplicate the images that contain hard classes (namely bicycle, chair, table, pottedplant, and sofa) in the training set. As shown in \figref{fig:bicycle}, the simple bootstrapping method is effective for segmenting the bicycle class. In the end, our `DeepLabv3' achieves the performance of 85.7\% on the test set without any DenseCRF post-processing, as shown in \tabref{tab:res_testset}.

\begin{table}[!t]
  \centering
  \addtolength{\tabcolsep}{2.5pt}
  \begin{tabular}{l | c}
    \toprule[0.2 em]
    {\bf Method} & {\bf mIOU} \\
    \toprule[0.2 em]
    Adelaide\_VeryDeep\_FCN\_VOC \cite{wu2016bridging} & 79.1 \\
    LRR\_4x\_ResNet-CRF \cite{ghiasi2016laplacian} & 79.3 \\
    DeepLabv2-CRF \cite{chen2016deeplab} & 79.7 \\
    CentraleSupelec Deep G-CRF \cite{chandra2016fast} & 80.2 \\
    HikSeg\_COCO \cite{sun2016mixed} & 81.4 \\
    SegModel \cite{shen2017semantic} & 81.8 \\
    Deep Layer Cascade (LC) \cite{li2017not} & 82.7 \\
    TuSimple \cite{wang2017understanding} & 83.1 \\
    Large\_Kernel\_Matters \cite{peng2017large} & 83.6 \\
    Multipath-RefineNet \cite{lin2016refinenet} & 84.2 \\
    ResNet-38\_MS\_COCO \cite{wu2016wider} & 84.9 \\
    PSPNet \cite{zhao2016pyramid} & 85.4 \\
    IDW-CNN \cite{wanglearning}  & 86.3 \\
    CASIA\_IVA\_SDN \cite{fu2017stacked} & 86.6 \\
    DIS \cite{luo2017deep} & 86.8 \\
    \midrule
    \href{http://host.robots.ox.ac.uk:8080/anonymous/EGZFFR.html}{DeepLabv3} & 85.7 \\
    \href{http://host.robots.ox.ac.uk:8080/anonymous/BDPJA4.html}{DeepLabv3-JFT} & 86.9 \\
    \bottomrule[0.1 em]
  \end{tabular}
  \caption{Performance on PASCAL VOC 2012 {\it test} set.}
  \label{tab:res_testset}
\end{table}

\textbf{Model pretrained on JFT-300M:} Motivated by the recent work of \cite{sun2017revisiting}, we further employ the ResNet-101 model which has been pretraind on both ImageNet and the JFT-300M dataset \cite{hinton2015distilling, chollet2016xception, sun2017revisiting}, resulting in a performance of 86.9\% on PASCAL VOC 2012 test set.

\section{Conclusion}
Our proposed model ``DeepLabv3'' employs atrous convolution with upsampled filters to extract dense feature maps and to capture long range context. Specifically, to encode multi-scale information, our proposed cascaded module gradually doubles the atrous rates while our proposed atrous spatial pyramid pooling module augmented with image-level features probes the features with filters at multiple sampling rates and effective field-of-views. Our experimental results show that the proposed model significantly improves over previous DeepLab versions and achieves comparable performance with other state-of-art models on the PASCAL VOC 2012 semantic image segmentation benchmark.

\ifcvprfinal
\paragraph{Acknowledgments}
We would like to acknowledge valuable discussions with Zbigniew Wojna, the help from Chen Sun and Andrew Howard, and the support from Google Mobile Vision team.
\fi

\appendix
\section{Effect of hyper-parameters}
In this section, we follow the same training protocol as in the main paper and experiment with the effect of some hyper-parameters.

\textbf{New training protocol:} As mentioned in the main paper, we change the training protocol in \cite{chen2014semantic, chen2016deeplab} with three main differences: (1) larger crop size, (2) upsampling logits during training, and (3) fine-tuning batch normalization. Here, we quantitatively measure the effect of the changes. As shown in \tabref{tab:crop_size}, DeepLabv3 attains the performance of 77.21\% on the PASCAL VOC 2012 \textit{val} set \cite{everingham2014pascal} when adopting the new training protocol setting as in the main paper. When training DeepLabv3 without fine-tuning the batch normalization, the performance drops to 75.95\%. If we do not upsample the logits during training (and instead downsample the groundtruths), the performance decreases to 76.01\%. Furthermore, if we employ smaller value of crop size (\ie, 321 as in \cite{chen2014semantic, chen2016deeplab}), the performance significantly decreases to 67.22\%, demonstrating that boundary effect resulted from small crop size hurts the performance of DeepLabv3 which employs large atrous rates in the Atrous Spatial Pyramid Pooling (ASPP) module.

\textbf{Varying batch size:} Since it is important to train DeepLabv3 with fine-tuning the batch normalization, we further experiment with the effect of different batch sizes. As shown in \tabref{tab:batch_size}, employing small batch size is inefficient to train the model, while using larger batch size leads to better performance.

\textbf{Output stride:} The value of $\emph{output\_stride}$ determines the output feature map resolution and in turn affects the largest batch size we could use during training. In \tabref{tab:output_stride}, we quantitatively measure the effect of employing different $\emph{output\_stride}$ values during both training and evaluation on the PASCAL VOC 2012 \textit{val} set. We first fix the evaluation $\emph{output\_stride}=16$, vary the training $\emph{output\_stride}$ and fit the largest possible batch size for all the settings (we are able to fit batch size 6, 16, and 24 for training $\emph{output\_stride}$ equal to 8, 16, and 32, respectively). As shown in the top rows of \tabref{tab:output_stride}, employing training $\emph{output\_stride}=8$ only attains the performance of 74.45\% because we could not fit large batch size in this setting which degrades the performance while fine-tuning the batch normalization parameters. When employing training $\emph{output\_stride}=32$, we could fit large batch size but we lose feature map details. On the other hand, employing training $\emph{output\_stride}=16$ strikes the best trade-off and leads to the best performance. In the bottom rows of \tabref{tab:output_stride}, we increase the evaluation $\emph{output\_stride}=8$. All settings improve the performance except the one where training $\emph{output\_stride}=32$. We hypothesize that we lose too much feature map details during training, and thus the model could not recover the details even when employing $\emph{output\_stride}=8$ during evaluation.

\begin{table}[!t]
  \centering
  \scalebox{1}{
  \begin{tabular}{c c c | c}
    \toprule[0.2em]
    Crop Size & UL & BN & mIOU \\
    \toprule[0.2em]
    513 & \checkmark & \checkmark & 77.21 \\
    513 & \checkmark &  & 75.95 \\
    513 &  & \checkmark & 76.01 \\
    321 &  & \checkmark & 67.22 \\
    \bottomrule[0.1em]
  \end{tabular}
  }
  \caption{Effect of hyper-parameters during training on PASCAL VOC 2012 \textit{val} set at $\emph{output\_stride=16}$. {\bf UL}: Upsampling Logits. {\bf BN}: Fine-tuning batch normalization.}
  \label{tab:crop_size}
\end{table}

\begin{table}[!t]
  \centering
  \scalebox{1}{
  \begin{tabular}{c | c}
    \toprule[0.2em]
    batch size & mIOU \\
    \toprule[0.2em]
    4 & 64.43 \\
    8 & 75.76 \\
    12 & 76.49 \\
    16 & 77.21 \\
    \bottomrule[0.1em]
  \end{tabular}
  }
  \caption{Effect of batch size on PASCAL VOC 2012 \textit{val} set. We employ $\emph{output\_stride=16}$ during both training and evaluation. Large batch size is required while training the model with fine-tuning the batch normalization parameters.}
  \label{tab:batch_size}
\end{table}

\begin{table}[!t]
  \centering
  \scalebox{0.9}{
  \begin{tabular}{c c | c}
    \toprule[0.2em]
    train $\emph{output\_stride}$ & eval $\emph{output\_stride}$ & mIOU  \\
    \toprule[0.2em]
    8 & 16 & 74.45 \\
    16 & 16 & 77.21 \\
    32 & 16 & 75.90 \\
    \toprule[0.1em]
    8 & 8 & 75.62 \\
    16 & 8 & 78.51 \\
    32 & 8 & 75.75 \\
    \bottomrule[0.1em]
  \end{tabular}
  }
  \caption{Effect of $\emph{output\_stride}$ on PASCAL VOC 2012 \textit{val} set. Employing $\emph{output\_stride=16}$ during training leads to better performance for both eval $\emph{output\_stride}=8$ and $16$.}
  \label{tab:output_stride}
\end{table}

\section{Asynchronous training}

In this section, we experiment DeepLabv3 with TensorFlow asynchronous training \cite{abadi2016tensorflow}. We measure the effect of training the model with multiple replicas on PASCAL VOC 2012 semantic segmentation dataset. Our baseline employs simply one replica and requires training time 3.65 days with a K80 GPU.
As shown in \tabref{tab:async_pascal}, we found that the performance of using multiple replicas does not drop compared to the baseline. However, training time with 32 replicas is significantly reduced to 2.74 hours.

\begin{table}[!t]
  \centering
  \scalebox{1}{
  \begin{tabular}{c | c c}
    \toprule[0.2em]
    num replicas & mIOU & relative training time \\
    \toprule[0.2em]
    1 & 77.21 & 1.00x \\
    2 & 77.15 & 0.50x \\
    4 & 76.79 & 0.25x \\
    8 & 77.02 & 0.13x \\
    16 & 77.18 & 0.06x \\
    32 & 76.69 & 0.03x \\
    \bottomrule[0.1em]
  \end{tabular}
  }
  \caption{Evaluation performance on PASCAL VOC 2012 \textit{val} set when adopting asynchronous training.}
  \label{tab:async_pascal}
\end{table}

\section{DeepLabv3 on Cityscapes dataset}
Cityscapes \cite{Cordts2016Cityscapes} is a large-scale dataset containing high quality pixel-level annotations of 5000 images (2975, 500, and 1525 for the training, validation, and test sets respectively) and about 20000 coarsely annotated images. Following the evaluation protocol \cite{Cordts2016Cityscapes}, 19 semantic labels are used for evaluation without considering the void label.

We first evaluate the proposed DeepLabv3 model on the validation set when training with only 2975 images (\ie, \textit{train\_fine} set). We adopt the same training protocol as before except that we employ 90K training iterations, crop size equal to 769, and running inference on the whole image, instead of on the overlapped regions as in \cite{chen2016deeplab}. As shown in \tabref{tab:cityscapes_val}, DeepLabv3 attains the performance of 77.23\% when evaluating at $\emph{output\_stride}=16$. Evaluating the model at $\emph{output\_stride}=8$ improves the performance to 77.82\%. When we employ multi-scale inputs (we could fit $scales=\{0.75, 1, 1.25\}$ on a K40 GPU) and add left-right flipped inputs, the model achieves 79.30\%.

\begin{figure*}[!t]
  \centering
  \begin{tabular}{c}
    \includegraphics[width=0.95\linewidth]{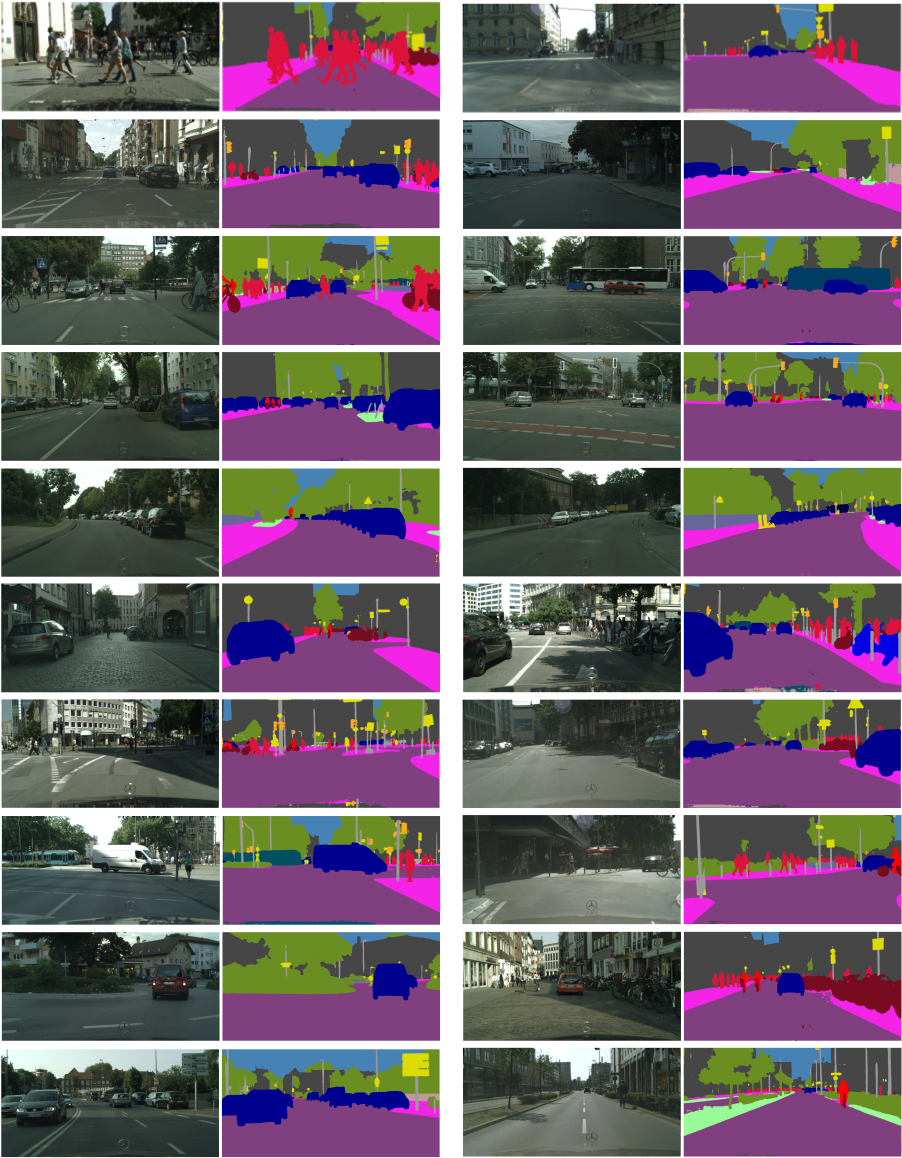} \\
  \end{tabular}
  \caption{Visualization results on Cityscapes {\it val} set when training with only {\it train\_fine} set.}
  \label{fig:vis_cityscapes}
\end{figure*}

In order to compete with other state-of-art models, we further train DeepLabv3 on the $\textit{trainval\_coarse}$ set (\ie, the 3475 finely annotated images and the extra 20000 coarsely annotated images). We adopt more scales and finer $\emph{output\_stride}$ during inference. In particular, we perform inference with $scales=\{0.75, 1, 1.25, 1.5, 1.75, 2\}$ and evaluation $\emph{output\_stride}=4$ with CPUs, which contributes extra 0.8\% and 0.1\% respectively on the validation set compared to using only three scales and $\emph{output\_stride}=8$. In the end, as shown in \tabref{tab:cityscapes_test}, our proposed DeepLabv3 achieves the performance of 81.3\% on the test set. Some results on \textit{val} set are visualized in \figref{fig:vis_cityscapes}.

\begin{table}[!t]
  \centering
  \scalebox{1}{
  \begin{tabular}{c c c c | c}
    \toprule[0.2em]
    OS=16            & OS=8       & MS         & Flip       & mIOU \\
    \toprule[0.2em]
    \checkmark      &            &             &             & 77.23 \\
                    & \checkmark &             &             & 77.82 \\
                    & \checkmark & \checkmark  &             & 79.06 \\
                    & \checkmark & \checkmark  & \checkmark  & 79.30 \\
    \bottomrule[0.1em]
  \end{tabular}
  }
  \caption{DeepLabv3 on the Cityscapes \textit{val} set when trained with only \textit{train\_fine} set. {\bf OS}: \emph{output\_stride}. {\bf MS}: Multi-scale inputs during inference. {\bf Flip}: Adding left-right flipped inputs.}
  \label{tab:cityscapes_val}
\end{table}

\begin{table}[!th]
  \centering
  \addtolength{\tabcolsep}{2.5pt}
  \begin{tabular}{l c | c}
    \toprule[0.2 em]
    {\bf Method} & {\bf Coarse} & {\bf mIOU} \\
    \toprule[0.2 em]
    DeepLabv2-CRF \cite{chen2016deeplab} & & 70.4 \\
    Deep Layer Cascade \cite{li2017not} & & 71.1 \\
    ML-CRNN \cite{fan2016multi} & & 71.2 \\
    Adelaide\_context \cite{lin2015efficient} & & 71.6 \\
    FRRN \cite{pohlen2016full} & & 71.8 \\
    LRR-4x \cite{ghiasi2016laplacian} & \checkmark & 71.8 \\
    RefineNet \cite{lin2016refinenet} & & 73.6 \\
    FoveaNet \cite{li2017fovea} & & 74.1 \\
    Ladder DenseNet \cite{kreso2017ladder} & & 74.3 \\
    PEARL \cite{jin2017video} & & 75.4 \\
    Global-Local-Refinement \cite{zhang2017global} & & 77.3 \\
    SAC\_multiple \cite{zhang2017scale} & & 78.1 \\
    SegModel \cite{shen2017semantic} & \checkmark & 79.2 \\
    TuSimple\_Coarse \cite{wang2017understanding} & \checkmark & 80.1 \\
    Netwarp \cite{gadde2017semantic} & \checkmark & 80.5 \\
    ResNet-38 \cite{wu2016wider} & \checkmark & 80.6 \\
    PSPNet \cite{zhao2016pyramid} & \checkmark & 81.2 \\
    \midrule
    DeepLabv3 & \checkmark & 81.3 \\
    \bottomrule[0.1 em]
  \end{tabular}
  \caption{Performance on Cityscapes {\it test} set. {\bf Coarse}: Use \textit{train\_extra} set (coarse annotations) as well. Only a few top models with known references are listed in this table.}
  \label{tab:cityscapes_test}
\end{table}


{\small
\bibliographystyle{ieee}
\bibliography{egbib}
}

\end{document}